\documentclass[letterpaper, 10 pt, conference]{ieeeconf}  % Comment this line out if you need a4paper

\IEEEoverridecommandlockouts                              % This command is only needed if 
\usepackage{graphics} % for pdf, bitmapped graphics files
\usepackage{graphicx}
\usepackage{multirow}
\usepackage{mathptmx} % assumes new font selection scheme installed
\usepackage{times} % assumes new font selection scheme installed
\usepackage{amsmath} % assumes amsmath package installed
\usepackage{amssymb}  % assumes amsmath package installed
\usepackage{siunitx}
\usepackage{blindtext}
\usepackage{booktabs}
\usepackage{xcolor}
\usepackage{tikz}
\usetikzlibrary{positioning}
\usetikzlibrary{arrows}
\usetikzlibrary{calc}
\usetikzlibrary{shapes.multipart}
\usetikzlibrary{arrows.meta}
\usetikzlibrary{shadings}

\title{\LARGE \bf
Improved Orientation Estimation and Detection with Hybrid Object Detection Networks for Automotive Radar
}%TODO

\author{Michael Ulrich$^{1}$, Sascha Braun$^{2}$, Daniel K\"ohler$^{2,3}$, Daniel Niederl\"ohner$^{2}$, Florian Faion$^{1}$, Claudius Gl\"aser$^{1}$ \\and Holger Blume$^{3}$% <-this % stops a space
\thanks{$^{1}$Michael Ulrich, Florian Faion and Claudius Gl\"aser are with Robert Bosch GmbH, Corporate Research, Germany
        {\tt\small michael.ulrich2@bosch.com}}% <-this % stops a space
\thanks{$^{2}$Sascha Braun, Daniel K\"ohler and Daniel Niederl\"ohner are with Robert Bosch GmbH, Cross-Domain Computing Solutions, Germany}%
\thanks{$^{3}$Daniel K\"ohler and Holger Blume are with Leibniz University Hannover, Germany}%
}

\begin{document}

\bstctlcite{IEEEexample:BSTcontrol} 

\def\cred{\textcolor{red}}
\def\cblue{\textcolor{blue}}
\def\cgreen{\textcolor{green}}

\newcommand\copyrighttextinitial{%
	
	\scriptsize This work has been submitted to the IEEE for possible publication. Copyright may be transferred without notice, after which this version may no longer be accessible.}%
\newcommand\copyrighttextfinal{%
	
	\scriptsize\copyright\ 2022 IEEE. Personal use of this material is permitted. Permission from IEEE must be obtained for all other uses, in any current or future media, including reprinting/republishing this material for advertising or promotional purposes, creating new collective works, for resale or redistribution to servers or lists, or reuse of any copyrighted component of this work in other works.}%
\newcommand\copyrightnotice{%
	
	\begin{tikzpicture}[remember picture,overlay]%
	
	\node[anchor=south,yshift=10pt] at (current page.south) {{\parbox{\dimexpr\textwidth-\fboxsep-\fboxrule\relax}{\copyrighttextfinal}}};%
	\end{tikzpicture}%
	
	%\vspace{-10pt}%
	
}

\newcommand{\ts}[1]{{\textsubscript{#1}}}
\newcommand{\tbf}[1]{{\textbf{#1}}}

\maketitle
\copyrightnotice
\thispagestyle{empty}
\pagestyle{empty}

%%%%%%%%%%%%%%%%%%%%%%%%%%%%%%%%%%%%%%%%%%%%%%%%%%%%%%%%%%%%%%%%%%%%%%%%%%%%%%%%
\begin{abstract}

This paper presents novel hybrid architectures that combine grid- and point-based processing to improve the detection performance and orientation estimation of radar-based object detection networks. 
Purely grid-based detection models operate on a bird's-eye-view (BEV) projection of the input point cloud.
These approaches suffer from a loss of detailed information through the discrete grid resolution. 
This applies in particular to radar object detection, where relatively coarse grid resolutions are commonly used to account for the sparsity of radar point clouds.
In contrast, point-based models are not affected by this problem as they process point clouds without discretization.
However, they generally exhibit worse detection performances than grid-based methods. 

We show that a point-based model can extract neighborhood features, leveraging the exact relative positions of points, before grid rendering. 
This has significant benefits for a subsequent grid-based convolutional detection backbone.
In experiments on the public nuScenes dataset our hybrid architecture achieves improvements in terms of detection performance ($19.7\%$ higher mAP for \textsl{car} class than next-best radar-only submission) and orientation estimates ($11.5\%$ relative orientation improvement) over networks from previous literature.
%TODO Quantify improvements (percentages), mention submission to official benchmark

%%This paper investigates, how good orientation estimates of oriented bounding boxes (OBBs) can be combined with a high detection performance. 
%This paper presents a novel method to improve orientation estimates of oriented bounding boxes of an object detection network in the context of automotive radar. 
%Radar object detection networks suffer from poor orientation estimates, when they operate on a bird's-eye-view (BEV) projection of the radar reflection point cloud. 
%This is due to the low resolution of automotive radar point clouds, in comparison to other sensors. 
%In contrast, continuous processing of point clouds improves this problem, at the cost of detection performance. 
%The general idea of this paper is to combine a continuous model operating on the radar reflection point cloud for good orientation estimates with a convolutional detector operating on the BEV for good detection performance. 
%The continuous model is able to learn orientation-related features, for example based on the exact relative position of points, before such information is lost when rendering the point cloud to a discrete BEV grid. 
%In experiments on the public nuScenes dataset we show an improvement over previous literature in terms of orientation estimates, while detection performance is slightly improved at the same time.
\end{abstract}

%%%%%%%%%%%%%%%%%%%%%%%%%%%%%%%%%%%%%%%%%%%%%%%%%%%%%%%%%%%%%%%%%%%%%%%%%%%%%%%%
\section{Introduction}

The tracking and classification of traffic participants is crucial for automated driving and advanced driver assistance systems. 
Object detection networks are often applied for this purpose in the lidar \cite{Shi2019,Yang2019} and computer vision community \cite{Redmon2016,He2017}. 
In the automotive context, radar sensors are commonly used for environment sensing, due to their low cost and robustness to e.g. weather and lighting. 
Today, radar perception is a helpful source of information for sensor fusion with camera and lidar, with increasing potential due to next-generation high resolution radar sensors. 
Recent radar perception research is increasingly investigating object detection networks \cite{Scheiner2021}. 
In that research, two competing directions can be observed: 
The first research direction is to down-scale object detection network architectures from the lidar- to the radar-domain \cite{Dreher2020, Xu2021, Niederloehner2022}. 
These methods usually render the lidar point cloud to a 2D BEV or 3D Cartesian grid and apply a convolutional neural network (CNN) on that representation. 
In contrast, the second research direction is to up-scale point-wise operating neural networks to object detection \cite{Danzer2019, Svenningsson2021}. 
While the grid-based CNN detectors are a mature technology adopted from computer vision, the point-based approaches proved to be suitable for radar data on related tasks, such as semantic segmentation \cite{Schumann2018}, object type classification \cite{Ulrich2021} or tracking \cite{Ebert2020}.

\begin{figure}[!tb]
	\centering
	\input{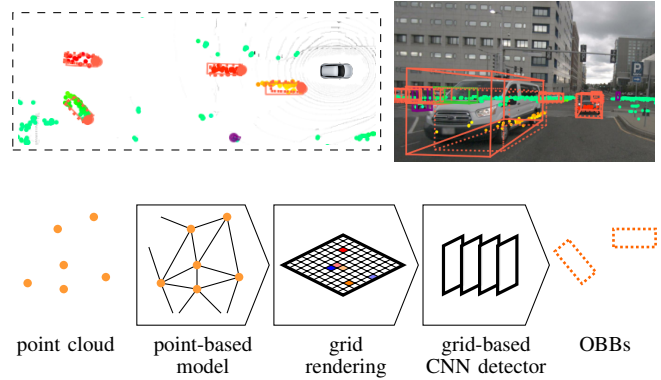}
	\caption{Radar-only object detection networks in the automotive context suffer from poor detection performance and orientation estimates. 
	This paper uses point-based layers, such as a graph neural network, to extract local neighborhood information in a first step. 
	The grid rendering allows the application of a grid-based, convolutional neural network (CNN) to detect oriented bounding boxes (OBBs). 
	By doing so, we combine the continuous processing of the point-based methods for good orientation estimates with the high detection performance of the grid-based methods.}
	\label{fig:teaser}
\end{figure}

Object detection networks produce oriented bounding boxes (OBBs), given a single or temporally aggregated measurement, which can be subsequently tracked, for example using a Kalman filter. 
The OBB is characterized by its position, orientation, shape (length and width) and class. 
In contrast, traditional radar perception performs object tracking first, followed by object type classification. 
The traditional tracking typically involves a handcrafted model, which explains how each object can produce multiple reflection points. 

Radar reflection lists are typically very sparse in comparison to lidar point clouds. 
Therefore, it is usually necessary to decrease the resolution of the BEV grid (larger grid cells), when applying grid-based object detection networks to radar. 
%Otherwise, most grid cells would not contain information from any point, and information is difficult to process with convolutional kernels. 
%However, a low grid resolution (e.g. $0.5$m cells for radar instead of $0.125$m cells for lidar) results in an information loss regarding the precise position and geometric structure of the reflections. 
%Therefore, the OBB parameters are difficult to estimate. 
An information loss of the precise position and geometric structure is the consequence, which can impede the prediction of shape and orientation of objects.
Deteriorated orientation estimates of OBBs are particularly problematic due to their importance for the subsequent tracking. 
% while point-based OD networks do not suffer from the loss of precise position information as they process the points individually.
% However, grid-based methods often provide a very good detection performance \cite{Scheiner2021, Xu2021, Nobis2021b, Lee2020} and our experiments showed that they outperform point-based methods.
In contrast, point-based object detection networks do not suffer from this problem, as they extract pointwise features and maintain the exact position of points.
However, grid-based methods often provide a very good detection performance \cite{Scheiner2021, Xu2021, Nobis2021b, Lee2020} and outperformed point-based models in this regard in our experiments.

In this work we propose a hybrid architecture for radar object detection, that utilizes a preceding point-based network to extract abstract pointwise features, before the points are rendered to a grid for further processing by a grid-based network.

The contributions of our paper are the following:
\begin{itemize}
	\item We investigate, how a point-based network can be used for feature extraction. 
	\item We combine point-based feature extraction with a grid-based object detection network.
	Thereby, we take the best of both worlds, achieving high detection performance and good orientation estimates at the same time. 
	\item We benchmark our algorithms on the public nuScenes dataset \cite{Caesar2020} and outperform reference algorithms from previous literature regarding detection performance and orientation estimation.
\end{itemize}

\section{Related Work}
Previous literature on radar object detection can be categorized by its input data. 
In the wake of deep learning, the trend to input data as raw as possible is often interpreted to provide radar spectra to the neural networks. 
While 1D radar velocity spectra or 2D (Doppler) spectrograms are commonly used for object type classification \cite{Ulrich2018a,Ulrich2018b}, radar spectra for object detection networks are usually 3D \cite{Palffy2020} (range, radial velocity, azimuth angle).
The spectra are often reduced to a single \cite{Brodeski2019,Wang2021a} or multiple \cite{Lim2019} 2D projections in order to apply 2D convolutional neural networks (CNNs). 
A 2D or 3D grid is a natural representation for spectrum data, hence most literature uses CNNs, with few exceptions, such as \cite{Meyer2021}. 

Other methods in the literature use the point cloud-like representation of reflection lists as input to the detection networks, see Tab.\,\ref{tab:literature_rod}. 
The reflections are a more abstract representation in comparison to spectra, which has the benefit of a higher robustness to sensor modifications, such as changes to the radar modulation. 
We group previous literature in this area into grid-based and point-based methods. 
Point-based object detection networks generate box proposals for a subset of the input point cloud, while grid-based detectors render the point cloud to a grid and generate box proposals for each grid cell. 
Further, we distinguish over which features of the reflection point cloud the essential feature aggregation is performed. 
For the grid-based approaches, the essential feature aggregation step is the grid rendering. 
Clearly, neighborhood context can still be exchanged e.g. by applying convolutional layers on the grid, but the context of points within the cells is already aggregated in an early stage. 
\cite{Dreher2020, Nobis2021b, Lee2020, Meyer2019a} project the point cloud directly to a grid. 
The handcrafted, point-wise features (such as radar cross section RCS or radial velocity $v_r$) are directly aggregated, for example using max- or mean-pooling, if multiple points fall into the same grid cell.
Alternatively, more abstract features of the points are learnt in architectures, which use a grid rendering module adopted from PointPillars \cite{Lang2019}, for example in \cite{Scheiner2021, Niederloehner2022, Palffy2022}. \cite{Xu2021} further extend the grid rendering by applying a self-attention mechanism to the cell-wise learnt features.

In contrast, point-based object detection networks omit the grid rendering step. 
Instead, \cite{Scheiner2020} use a parametric clustering algorithm, followed by the extraction of handcrafted features. 
\cite{Danzer2019, Bansal2020} use a grouping (pooling) mechanism adopted from PointNet/PointNet++ to aggregate features of a neighborhood hierarchically.
Information sharing of points in a local neighborhood is achievable by message passing (graph neural network, GNN) \cite{Svenningsson2021}, 
kernel point convolutions (KPConv) \cite{Nobis2021a} or an attention mechanism (transformer) \cite{Bai2021}. 
Our work closes the gap of combining feature-learning of point neighborhoods with a grid-based radar object detection network, cf. Tab.\,\ref{tab:literature_rod}. 

\begin{table}
	\centering
	\caption{Categorization of radar object detection methods, which use the radar reflection list (point cloud).
	}
	\label{tab:literature_rod}
	\begin{tabular}{c c c}
		\toprule
		\begin{tabular}{@{}c@{}}\textbf{Essential feature} \\ \textbf{aggregation over}\end{tabular} &
		\begin{tabular}{@{}c@{}}\textbf{Point-based} \\ \textbf{aggregation over}\end{tabular} & \begin{tabular}{@{}c@{}}\textbf{Grid-based} \\ \textbf{rendering over}\end{tabular} \\
		\hline 
		\begin{tabular}{@{}c@{}}point-wise handcrafted \\ features\end{tabular} & \cite{Scheiner2020} & \cite{Dreher2020, Nobis2021b, Lee2020, Meyer2019a}\vspace{0.1cm}\\
		point-wise learned features & \cite{Danzer2019, Bansal2020} & \cite{Scheiner2021,Xu2021,Niederloehner2022, Palffy2022}\vspace{0.1cm}\\
		\begin{tabular}{@{}c@{}}learned features of points \\ and neighborhood\end{tabular}  & \cite{Svenningsson2021, Nobis2021a, Bai2021} & this work \\
		\bottomrule
	\end{tabular}
\end{table}

\begin{figure*}[!tb]
	\centering
	\input{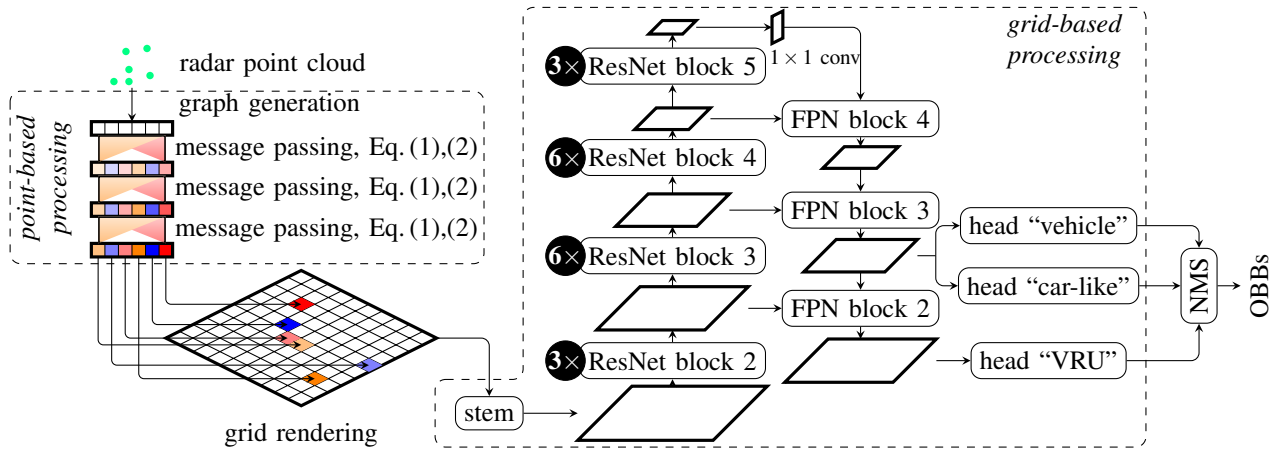}
	\caption{Architecture of the hybrid point-based and grid-based GraphPillars. 
		The local context of points relative to each other can already be learned in the point-based processing, before the coarse resolution of the grid discretizes the positions. 
		After grid rendering, a convolution backbone is used to extract feature maps using a ResNet and Feature Pyramid Network (FPN) like architecture. Convolutional detection heads for different groups of classes predict OBB proposals. Non-maximum suppression (NMS) is applied as a post-processing step. 
		The KPConvPillars architecture is analogous, but uses Kernel Point Convolutions Eq.\,(\ref{eq:kpconv_distance}),\,(\ref{eq:kpconv_weight}) instead of the point-based message passing. }
	\label{fig:network_architecture}
\end{figure*}
\section{Problem summary}
\textbf{Why are detection and orientation estimates important?}
The concept of tracking-by-detection is to extract OBBs from measurements first. 
The trackers fundamental information source are the OBBs, the raw radar sensor data is not used. 
Therefore, a high detection rate with few false-positives is desired. 
In the automated driving context, the object state vector often contains the heading angle, particularly when tracking vehicles. 
Therefore, a direct input of orientation (heading) of the OBB is very helpful. 
Furthermore, orientation estimates facilitate the object spawning significantly because the region of interest to verify candidates in subsequent measurements can be limited to the possible driving direction. 

\textbf{What is the bottleneck of current grid-based detectors?}
Grid-based detectors render the radar reflection point cloud to a discretized representation (typically in BEV) early in the processing. 
Detailed information of the points positions and therefore the object shape and orientation is lost, when the grid is relatively coarse. 
However, such coarse grids are necessary in radar object detection, due to the low resolution of the radar sensors. 

\textbf{Why do we still want to use grid-based detectors?}
On the one hand, grid-based object detection networks suffer from the information loss of the grid rendering.
But on the other hand, their detection performance is very high, as will be shown in Sec.\,\ref{sec:experiments}. 
A possible explanation could be the relatively mature technology of 2D CNNs, which is well investigated in similar fields, such as lidar object detection. 
%In contrast, only little research exists on point-based methods, such as GNNs, KPConvs and attention. 

\section{Method}
\subsection{Overview}
The general idea of the method is to combine point-based and grid-based models into a hybrid network. 
A preceding point-processing network of multiple layers is utilized to transform the features of the input point cloud, leveraging the exact position and geometric structure of points.

After these layers, each point holds an abstract feature vector, which encodes the local context as well as additional features such as the radial velocity $v_r$ or the RCS. 
Subsequently, these abstract features are rendered to a BEV grid and further processed by a grid-based object detection network.

In this work, we consider two realizations of the proposed hybrid architecture using different pointwise feature encoders.
The first architecture, referred to as GraphPillars, utilizes a graph neural network (GNN) which aggregates information from a graph of spatially proximal points.
The second model, which we call KPConvPillars, is equal to the GraphPillars network, except that it uses kernel point convolutions (KPConvs) instead of a GNN to extract features from the local neighborhood of points.

In contrast to previous grid-based networks, the features of the points already include their local neighborhood context, when they are rendered to the grid in our hybrid architecture. 
This facilitates the embedding of local shape and exact, relative positions, which is helpful to learn the OBB orientation and improves detection performance. 

\subsection{Point Feature Extractor and Grid Rendering}
\label{sec:point_feature_extractor}
\textbf{Graph Neural Network}
In this paper, we investigate two variants of point feature extractors. 
The first variant is a graph neural network (GNN), similar to \cite{Svenningsson2021,Shi2020}. 
The basic idea is to connect points in a local neighborhood (e.g.  \SI{1}{m} or \SI{2}{m}) by directed edges to form a graph. 
\begin{align}
	f_{\text{m}}(l,k) &= g_l(f_{\text{n}}(l,k), f_{\text{e}}(k)) \label{eq:message_embedding}\\
	f_{\text{n}}(l+1) &= f_{\text{n}}(l) + \rho([f_{\text{m}}(l,k)]_{N_n}) \label{eq:graph_feature}
\end{align}
In each layer, features (so-called messages $f_{\text{m}}$) are generated for each edge $k$ in layer $l$.
The message is a learnt encoding of the features of the sender (edge origin) node $f_{\text{n}}$ and some pre-defined feature of the edge $f_{\text{e}}$, cf. Eq.\,\ref{eq:message_embedding}. 
For example, $f_{\text{e}}(k)$ could be the relative position and relative $v_r$ of the two points. 
$g_l(\cdot)$ is a trainable function, in our case a multilayer perceptron consisting of $3$ fully-connected layers with ReLU activation. 
The same $g_l(\cdot)$ is applied to generate the embedding of all messages (weight sharing) in layer $l$. 
Finally, all messages $[\cdot]_{N_n}$ from the local neighborhood $N_n$, which are received by node $n$, are pooled using the pooling function $\rho(\cdot)$ (e.g. max-pooling) to generate a point-wise feature $f_{\text{n}}(l+1)$. 
In our experiments, we noted that the skip-connection in Eq.\,\ref{eq:graph_feature}, which adds the features of the previous layer, helps convergence significantly. 
The GNN consists of a sequence of such message passing layers.  

\begin{table*}[t]
	\centering
	\caption{Benchmark for class \textit{car} of point-based, grid-based and hybrid radar object detection networks on the nuScenes validation dataset and relative improvements compared to our baseline (PointPillars-like).
	}
	\label{tab:benchmark_nuscenes}
	\begin{tabular}{l c c c c c c c}
		\toprule
		\textbf{Object detector} & \textbf{Type} & \textbf{AP4.0 (\%) $\uparrow$} & \textbf{mAP (\%) $\uparrow$} & \textbf{AOE (rad) $\downarrow$} & \textbf{rel. AP4.0 $\uparrow$} & \textbf{rel. mAP $\uparrow$} & \textbf{rel. AOE $\downarrow$} \\
		\hline
		GNN \cite{Svenningsson2021}-like & point-based & 24.7 & 13.7 & 0.435 & -\SI{33.2}{\%} & -\SI{37.7}{\%} & +\SI{37.2}{\%} \\
		KPConv \cite{Thomas2019}-like & point-based & 28.5 & 15.7 & 0.615 & -\SI{23.0}{\%} & -\SI{28.6}{\%} & +\SI{94.0}{\%}\vspace{.1cm} \\
		BEV-rendering \cite{Yang2019}-like & grid-based & 36.7 & 20.6 & 0.391 & -\SI{0.8}{\%} & -\SI{6.4}{\%} & +\SI{23.3}{\%} \\
		PointPillars \cite{Lang2019}-like & grid-based & 37.0 & 22.0 & 0.317 & baseline & baseline & baseline \\
		RPFA-Net \cite{Xu2021}-like & grid-based & 38.3 & 23.1 & 0.320 & +\SI{3.5}{\%} & +\SI{+5.0}{\%} & +\SI{0.9}{\%}\vspace{.1cm} \\
		GraphPillars (ours) & hybrid & 40.2 & 24.6 & \textbf{0.286} & +\SI{8.6}{\%} & +\SI{11.8}{\%} & \textbf{-\SI{9.8}{\%}} \\
		KPConvPillars (ours) & hybrid & \textbf{42.2} & \textbf{26.2} &  0.321 & \textbf{+\SI{14.1}{\%}} & \textbf{+\SI{19.1}{\%}} & +\SI{1.3}{\%}\\
		\bottomrule
	\end{tabular}
\end{table*}

\textbf{Kernel Point Convolution}
The second point-based feature extractor considered in this work is the rigid kernel point convolution (KPConv) analogous to \cite{Thomas2019,Nobis2021a}. 
The basic idea of KPConvs is to extract features from the local neighborhood $N_x$ of a point $x$ using a set of pre-defined kernel points $x_k$ that carry learnable convolution weights $W_k$.
\begin{align}
h(x_k, y) = \max \left(0, 1 - \frac{\|x_k - y\|}{\sigma}\right) \label{eq:kpconv_distance}\\
f_{x}(l+1) = \sum_{x_i \in N_x} f_{x_i}(l) \sum_{k} h(x_k, x_i - x) W_k\label{eq:kpconv_weight}
\end{align}
A distance-based weighting function $h(\cdot)$ is used to determine the influence of each kernel point $x_k$ on a given point $y$, where the parameter $\sigma$ is a hyperparameter to define the neighborhood size, see Eq.\,\ref{eq:kpconv_distance}. 
The features $f_{x_i}(l)$ of input points at layer $l$ are transformed by multiplying them with the weights of each kernel point in Eq.\,\ref{eq:kpconv_weight}. 
The output feature $f_{x}(l+1)$ at coordinate $x$ is obtained by summing over the transformed features of all kernel points in the local neighborhood. 
In this work, we utilize KPConv layers with the structure of a ResNet block \cite{He2016}, where the 2D convolution is replaced with a KPConv, to obtain a point-based feature extractor as proposed by \cite{Thomas2019}. As for the GNN, a sequence of such layers forms the preceding point-processing network for our KPConvPillars architecture.

\textbf{Grid Rendering}
The grid rendering following the point-processing network is analogous to previous grid-based radar object detection methods and adopts the idea of PointPillars \cite{Lang2019}. 
The points are projected onto a Cartesian BEV grid, whose cells contain the features of the points. 
When multiple points fall into the same cell, the feature vectors of these points are aggregated, e.g. using mean-pooling. 
Consistently with \cite{Scheiner2021}, we found empirically that a grid cell size of $0.5\times 0.5$ meter achieves a good detection performance.

\subsection{Backbone and Detection Heads}
\label{sec:backbone}
We found empirically that the backbone proposed by the lidar PointPillars \cite{Lang2019} does not work well for radar, as also noted in \cite{Scheiner2021}. 
Instead we adopt the lightweight approach from \cite{Yang2019} that uses a 2D convolutional backbone consisting of a convolutional stem, a residual network \cite{He2016} and a feature pyramid network \cite{Lin2017} to process the BEV feature map at different resolutions, see Fig.\,\ref{fig:network_architecture}. 
For a detailed description of the backbone we refer to \cite{Yang2019}. 
%The residual network \cite{He2016} utilizes 2D convolutions and consists of $4$ stages and a stem, where the stem contains two convolutional layers with kernel size $3\times3$, separated by batch-norm and ReLU activation. 
%Each residual stage consists of $3$, $6$, $6$ and $3$ residual blocks, respectively. 
%A residual block has a $1\times1$ convolution, a $3\times3$ convolution and a $1\times1$ convolution (all followed by batch-norm and ReLU activation) on the main path, and a $1\times1$ convolution and batch-norm on the skip-connection path.
%The $1\times1$ convolution on the skip path is only used in the first block of each stage, which achieves a downsampling of the grid by factor $2$ through a stride. 
%After the bottleneck (a $1\times1$ 2D convolution), there is an up-scaling of the grid by a factor of $2$ at each stage, such that abstract features are obtained at different scales, similar to the feature pyramid network (FPN) \cite{Lin2017}. 
%This is achieved by a transposed $3\times3$ 2D convolution in the top-down path and a $1\times1$ 2D convolution in the lateral skip-connection path to the feature map of the corresponding resolution of the residual network, before the two convolution results are added. 

To predict the final OBBs, we use separate, fully convolutional (CNN) detection heads for different groups of classes. 
For example, the classes bus, truck, pickup and car trailer are combined in one head, which is attached to a feature map of a relatively low resolution. 
In contrast, the detection head for vulnerable road users (VRUs) is connected to a feature map of a finer grid resolution. 
Although each head predicts multiple classes (e.g. pedestrians and bicycles), we believe that the learning of certain regression parameters such as length, width and orientation differs significantly for e.g. VRUs and large vehicles, which motivates separate feature maps in the heads. 
The heads predict a class score (one-hot encoding) for each BEV grid cell, along with regression parameters for the OBBs such as length or width. 
The orientation is encoded as the sine and cosine of the orientation angle, to avoid discontinuities due to the $2\pi$ ambiguity of angles, as in \cite{Yang2019}. 
Further regression parameters are the $x$- and $y$-offset of the OBB center relative to the BEV grid cell. Consequently, all networks considered in this work are one-stage object detectors. 

\subsection{Training of the Hybrid Model}
For our two hybrid architectures, we use three message passing layers or KPConv layers respectively as point-based feature extractors before rendering the features to the BEV grid.
These models are trainable end-to-end. 
%However, we noted that a pre-training of the point-based layers improves detection performance by a few percent. 
%For this purpose, a point-based object detection network is trained to solve the object detection task.
%For example, a GNN consisting of $8$ graph message passing layers as described in Eq.\,\ref{eq:message_embedding},\,\ref{eq:graph_feature} is combined with a detection head consisting of $2$ fully connected layers and $1$ fully connected layer for classification and regression parameters each. 
%After training the point-based network, its first $3$ layers are frozen and used as a feature extractor in the hybrid model. 
%The class scores are trained using a focal loss while the box regression parameters are trained using an $l1$ loss. 
The classification loss (a focal loss) is weighted by a factor between $10$ (cars) and $200$ (VRUs) in comparison to the $L1$ regression loss. 
Furthermore, data augmentation is helpful, for example rotation of the whole point cloud by a small, random angle, random shifts, or horizontal flipping.

\section{Experiments}
\label{sec:experiments}
We evaluate on the nuScenes benchmark dataset, which consists of real-world recordings of driving situations in different countries. 
For comparability, the official nuScenes metrics \cite{Caesar2020} are used, unlike many other radar object detection literature. 
In particular, there is no filtering of OBBs applied based on field-of-view, the objects' movement state (stationary or dynamic), or the requirement of at least one radar point in the OBB to be detected. 
Clearly, this makes it more difficult to achieve high performance values, but makes our research comparable, also with respect to other modalities. 

%\begin{table}
%	\centering
%	\caption{Benchmark of point-based, grid-based and hybrid radar object detection networks on nuScenes dataset.
%	}
%	\label{tab:benchmark_nuscenes}
%	\begin{tabular}{l c c c}
%		\toprule
%		\textbf{Object detector} & \textbf{AP$\mathbf{4.0}$ $[\%]$ $\uparrow$} & \textbf{mAP $[\%]$ $\uparrow$}  & \textbf{AOE $[\text{rad}]$ $\downarrow$} \\
%		\hline 
%		GNN & 21.5& 11.7 & 0.446\\
%		KPConv & 25.3 & 13.7 & 0.625\vspace{0.1cm}\\
%		BEV-rendering & 33.3 & 18.2 & 0.390\\
%		PointPillars-like & 33.5 & 19.3 & 0.322\vspace{0.1cm}\\
%		%KPConvBEV (ours) & 35.7 & 20.8 & 0.308 \\
%		GraphPillars (ours) & 36.4 & 21.7 & \textbf{0.283} \\
%		KPConvPillars (ours) & \textbf{38.5} & \textbf{23.2} & 0.328\\
%		% RPFA-Net & & & \\
%		\bottomrule
%	\end{tabular}
%\end{table}
%{\setlength{\tabcolsep}{.45em}
%\begin{table}
%	\centering
%	\caption{Relative improvement of orientation estimates.
%	}
%	\label{tab:orientation_nuscenes}
%	\begin{tabular}{l c c c}
%		\toprule
%		\textbf{Object detector} & \textbf{AP$\mathbf{4.0}$ $[\%]$ $\uparrow$} & \textbf{AOE $[\text{rad}]$ $\downarrow$} &  \textbf{rel. AOE $\downarrow$}\\
%		\hline 
%		PointPillars-like & 33.5 & 0.322 & baseline \\
%		RPFA-Net-like & 34.6 & 0.320 & -\SI{0.7}{\%}\\
%		KPConvBEV (ours) & 35.7 & 0.308 & -\SI{4.3}{\%}\\
%		GraphPillars (ours) & \textbf{36.4} & \textbf{0.283} & \textbf{-\SI{12.1}{\%}} \\
%		\bottomrule
%	\end{tabular}
%\end{table}
%}

The nuScenes detection benchmark provides average precision metrics for a set of different matching thresholds $\left\{0.5, 1, 2, 4\right\}$\,\si{meters} as well as the mean average precision (mAP) for these thresholds to evaluate the detection performance of models.
The average precision is high, when many actual objects are detected and few false-positive (ghost) objects occur. 
For our evaluation we focus on the average precision for a matching threshold of \SI{4}{m} (AP4.0) and the mAP (higher AP4.0/mAP is better).
In this context, the mAP refers to the average of AP$4.0$, AP$2.0$, AP$1.0$ and AP$0.5$ for a specific class, rather than the average over all classes.
Furthermore, the nuScenes metrics include the average orientation error (AOE) to assess the performance regarding the orientation estimation.
The orientation error (OE) is the angle difference in radians between a ground-truth OBB and a detected OBB, which are matched with a threshold of \SI{2}{meters} center distance. 
The AOE is the cumulative mean of the OE over recall curve (different objectness thresholds for the detection OBB), where values below \SI{10}{\%} are clipped (lower AOE is better). 
The AOE is a true-positive metric, meaning that it does not consider false-positives or false-negatives. 

In our quantitative evaluation, we consider the above metrics only for class ``car'', as in previous radar object detection literature. 
Our network can for example detect objects of type ``bus'' or ``truck'' as well, however, the two classes are often confused with each other, cf. Fig.\,\ref{fig:exemplary_results}, scenes 1 and 2. 
This leads to low AP$4.0$/mAP for these classes, despite reasonable detections. 
%Further, the precise localization of large vehicles is difficult from radar data only, which has a high impact on the center-distance based matching metrics of nuScenes. 

In our comparison, we evaluate two point-based methods GNN and KPConv, three grid-based methods with conventional BEV-rendering, PointPillars grid rendering and the self-attention mechanism proposed by \cite{Xu2021} (RPFA-Net-like), as well as our two hybrid methods GraphPillars and KPConvPillars, which render learnt, local neighborhood features to the grid. 
\begin{itemize}
	\item The GNN is our re-implementation of \cite{Svenningsson2021} (8 message passing layers), where we note that the performance is improved to the original paper (our AP$4.0$ is \SI{3.1}{\%} better and our mAP is \SI{2.2}{\%} better). One explanation for this difference might be our data augmentation methods and the different loss function.
	\item KPConv is our re-implementation of \cite{Thomas2019} (12 KPConv layers), which processes the input point cloud at varying densities using a subsampling strategy. 
	\cite{Nobis2021a} evaluated a radar-based network using KPConvs on nuScenes, but for semantic segmentation and hence not with the original detection benchmark metrics. 
	\item The grid-based BEV-rendering is our re-implementation of the lidar model in \cite{Yang2019}. The radar-specific differences are the above mentioned grid resolution of $0.5\times 0.5$ meter, only a single height channel due to the missing elevation information, and one channel for each radar feature ($v_r$, RCS). 
	\item The PointPillars-like architecture uses the PointPillars module \cite{Lang2019} to render the point cloud to a BEV grid, but equals the BEV-rendering model otherwise, for comparability. 
	\item The RPFA-Net-like model uses our re-implementation of the feature attention module from \cite{Xu2021}, which follows the grid rendering in the otherwise equal PointPillars-like model. The experiments in \cite{Xu2021} were not performed on the nuScenes dataset. 
	\item The GraphPillars model is the proposed hybrid architecture of Fig.\,\ref{fig:network_architecture}, where three message passing (GNN) layers are used for the point-based preprocessing. 
	\item KPConvPillars equals the GraphPillars model, except that the message passing layers in the preceding point-based network are replaced by KPConv layers. 
\end{itemize}

\begin{table}[!t]
	\centering
	\caption{Test dataset results from the official nuScenes submission server: Radar-only object detection for class \textit{car}.
	}
	\label{tab:nuscenes_testset}
	\begin{tabular}{l c c c c}
		\toprule
		\textbf{Object detector} &  \textbf{mAP (\%) $\uparrow$} & \textbf{ATE $\downarrow$} & \textbf{ASE $\downarrow$} & \textbf{AOE $\downarrow$}\\
		\hline
		Radar-PointGNN \cite{Svenningsson2021} & 10.1 & 0.69 & 0.20 & 0.38\\
 		KPConvPillars (ours) & \textbf{29.8} & \textbf{0.589} & \textbf{0.183} & \textbf{0.336} \\
		\bottomrule
	\end{tabular}
\end{table}

NuScenes allows only few submissions to be evaluated on the test dataset, hence our in-depth comparison is performed on the official validation dataset. 
Quantitative results are summarized in Tab.\,\ref{tab:benchmark_nuscenes}. 
It can be observed, that the detection performance of the grid-based approaches is better than for the point-based networks. 
Also, the AOE of the point-based networks is worse than the AOE of the grid-based detectors.
Furthermore, we observe that the feature attention module (RPFA-Net-like) from \cite{Xu2021} slightly improves the mAP and deteriorates the AOE, in comparison to the PointPillars-like model.
We believe the reason for this is that the self-attention mechanism is applied after rendering the points to pillars, where the relative position of points from different pillars is already lost. 

%However, a direct comparison of the AOEs is difficult because the AOE is a true-positive metric, which considers only matched detection and ground truth OBB. 
Comparing the mAP for grid-based and hybrid detectors shows that a relative improvement of \SI{11.8}{\%} for GraphPillars and \SI{19.1}{\%} for KPConvPillars is possible through the point-based preprocessing.
We note that the improvement in terms of mAP (rel. mAP) is larger than the improvement in terms of AP$4.0$ (rel. AP$4.0$), which means that the performance gain is larger for smaller matching distances. 
The detector is able to locate the objects more precisely. 
This indicates that the proposed point-based preprocessing indeed helps to overcome the limitations of the discrete, coarse grid resolution. 

Furthermore, the GraphPillars architecture exhibits an orientation improvement of \SI{9.8}{\%} in AOE w.r.t. to the baseline.
Note that the majority of objects in the nuScenes dataset are oriented longitudinally with respect to the ego vehicle. 
We observed a strong bias in the estimates towards such orientations when the network cannot unambiguously determine the actual value from the input data. 
Since the AOE averages over all true positive detections and the orientation distribution is highly imbalanced, the network can already achieve a good AOE performance by predicting an orientation close to the bias.
Thus, a decrease of $0.031$ \si{rad} in AOE compared to the baseline is quite significant, considering that only a small amount of samples contribute to the improvement.
Some examples are shown in Fig.\,\ref{fig:exemplary_results}, where our hybrid approaches clearly improve the orientation of objects that are not longitudinal, while maintaining the good detection performance of a grid-based approach. 

Results on the official nuScenes test dataset are shown in Fig.\,\ref{tab:nuscenes_testset}.
The numbers are obtained from the benchmark server, where we submitted our results of the KPConvPillars model. 
The average precision is even better than on the validation dataset (the AP$4.0$ is \SI{48.8}{\%}) for the \textit{car} class and the AOE is worse. 
A possible explanation could be the different scenes in validation and test datasets. 
Furthermore, our KPConvPillars outperforms the only other radar-only official submission \cite{Svenningsson2021} to the nuScenes benchmark in terms of mAP as well as average translation error (ATE), average scale error (ASE) and average orientation error (AOE). 

\newcommand{\picturewithcoordinates}[5]{
	\begin{tikzpicture}
		\node (image) at (0,0) {\includegraphics[height=0.1\paperheight]{#1}};
		\draw [red,-{Stealth}, line width=1pt] #2--#3; 
		\draw [blue,-{Stealth}, line width=1pt] #4--#5; 
	\end{tikzpicture}
}

\begin{figure*}[!p]
	\centering
	\begin{tabular}{c c c c}
		& \textbf{Scene 1} & \textbf{Scene 2} & \textbf{Scene 3}\\
		\raisebox{1.2\totalheight}{\rotatebox[origin=c]{90}{Camera}}
		& \picturewithcoordinates{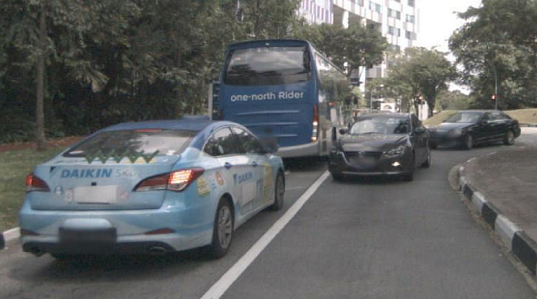}{(0, -1.35)}{(0, -0.85)}{(0, -1.35)}{(1.0, -1.35)}
		& \picturewithcoordinates{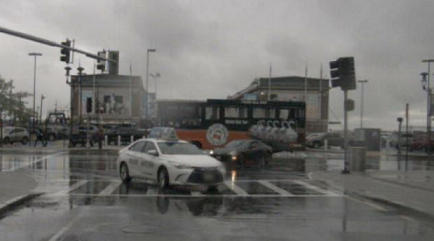}{(0, -1.35)}{(0, -0.85)}{(0, -1.35)}{(1.0, -1.35)}
		& \picturewithcoordinates{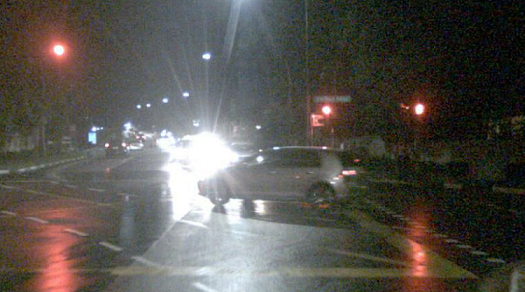}{(0, -1.35)}{(0, -0.85)}{(0, -1.35)}{(1.0, -1.35)}
		\\
		\raisebox{1.9\totalheight}{\rotatebox[origin=c]{90}{GNN}} 
		& \picturewithcoordinates{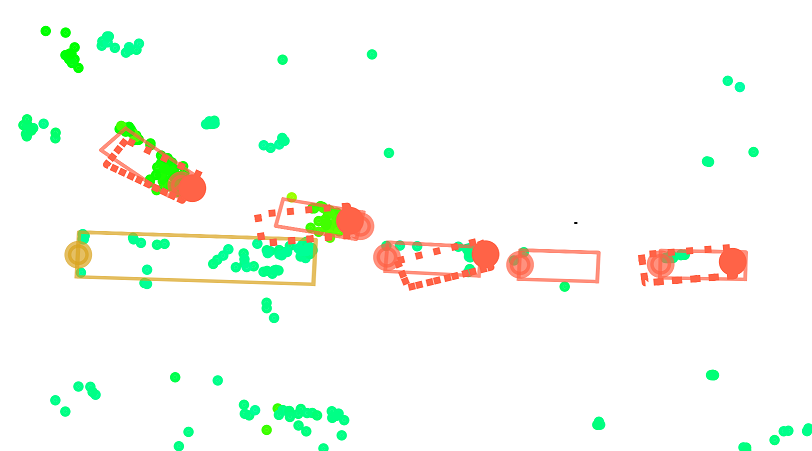}{(2.5, 0)}{(2.0, 0)}{(2.5, 0)}{(2.5, 1.0)}
		& \picturewithcoordinates{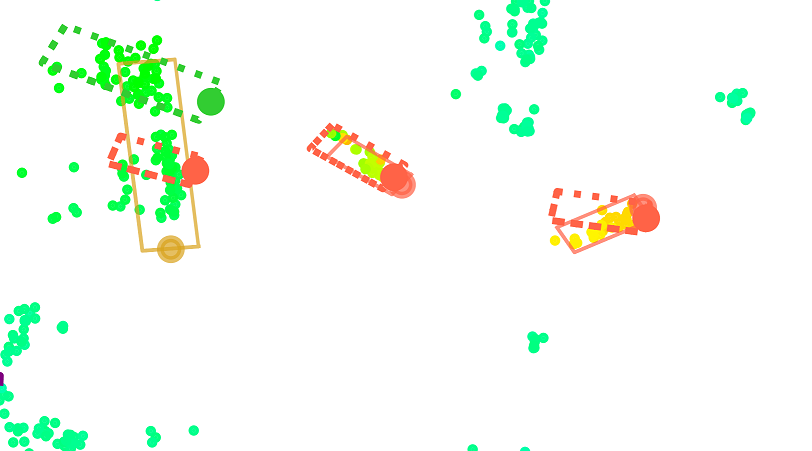}{(2.5, 0)}{(2.0, 0)}{(2.5, 0)}{(2.5, 1.0)}
		& \picturewithcoordinates{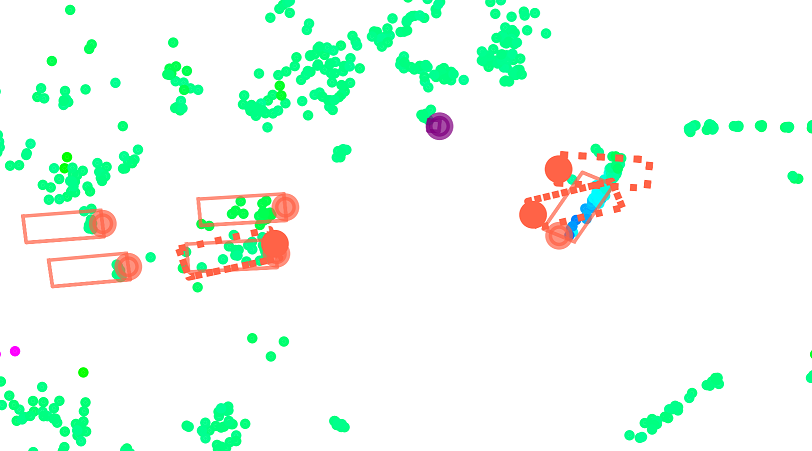}{(2.5, 0)}{(2.0, 0)}{(2.5, 0)}{(2.5, 1.0)}
		\\
		\raisebox{1.2\totalheight}{\rotatebox[origin=c]{90}{KPConv}}
		& \picturewithcoordinates{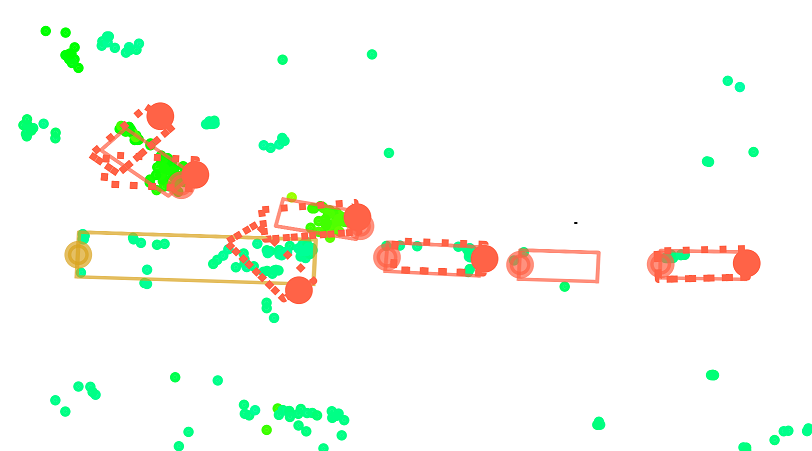}{(2.5, 0)}{(2.0, 0)}{(2.5, 0)}{(2.5, 1.0)}
		& \picturewithcoordinates{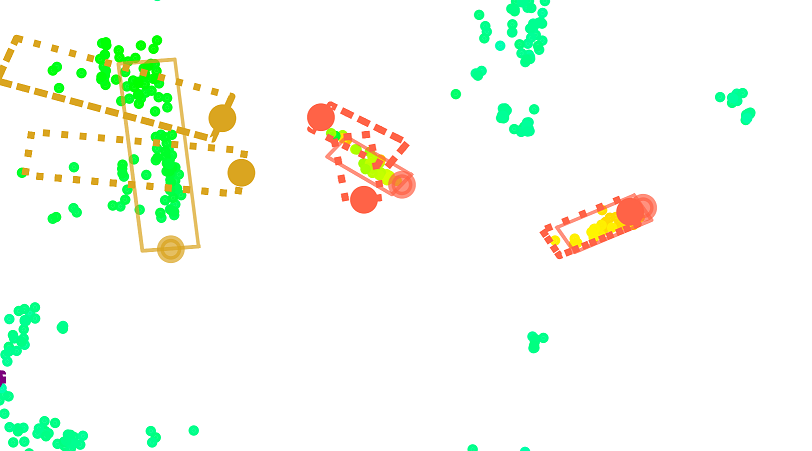}{(2.5, 0)}{(2.0, 0)}{(2.5, 0)}{(2.5, 1.0)}
		& \picturewithcoordinates{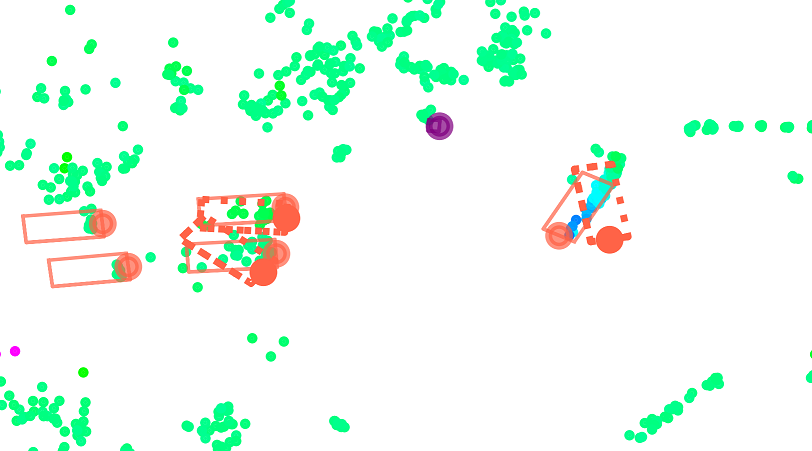}{(2.5, 0)}{(2.0, 0)}{(2.5, 0)}{(2.5, 1.0)}
		\\
		\raisebox{0.6\totalheight}{\rotatebox[origin=c]{90}{PointPillars-like}}
		& \picturewithcoordinates{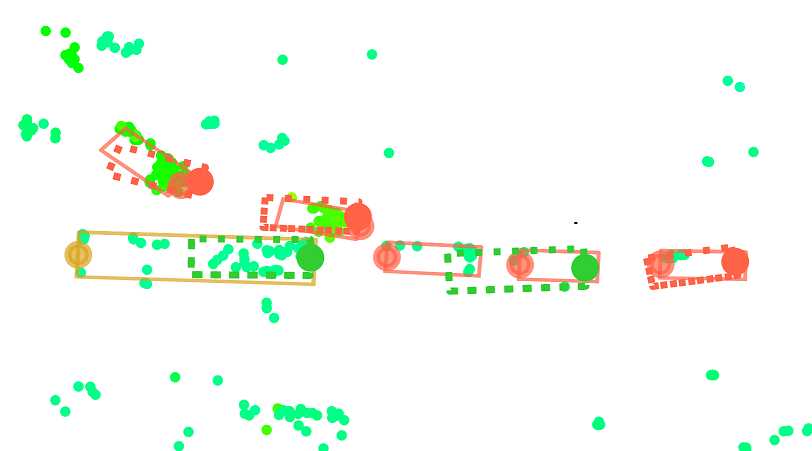}{(2.5, 0)}{(2.0, 0)}{(2.5, 0)}{(2.5, 1.0)}
		& \picturewithcoordinates{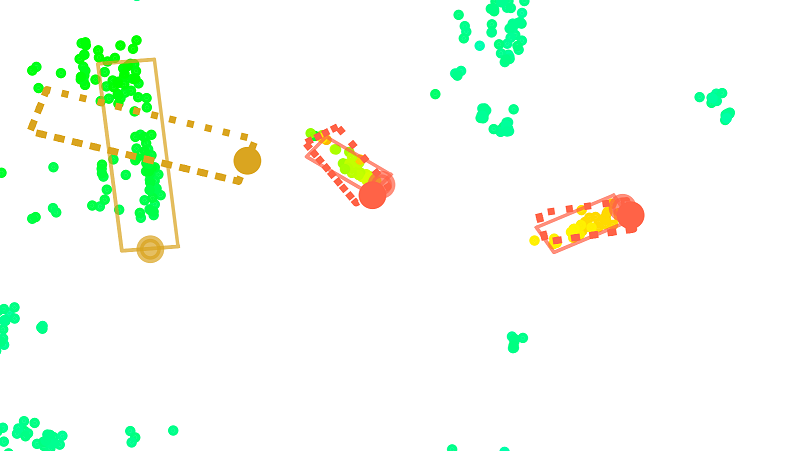}{(2.5, 0)}{(2.0, 0)}{(2.5, 0)}{(2.5, 1.0)}
		& \picturewithcoordinates{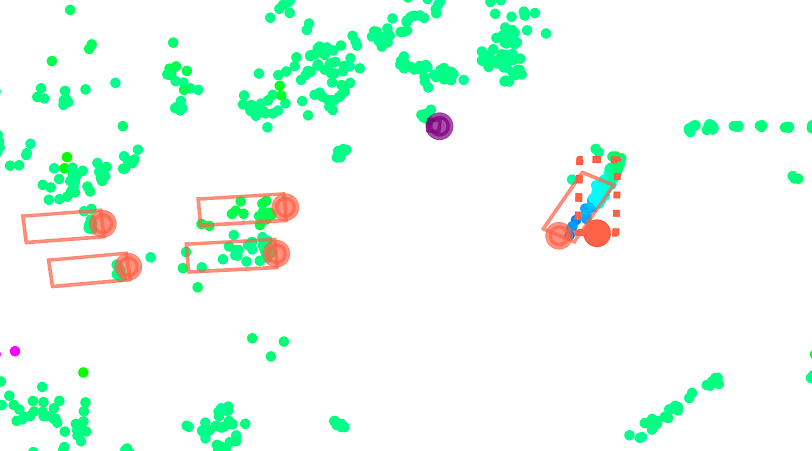}{(2.5, 0)}{(2.0, 0)}{(2.5, 0)}{(2.5, 1.0)}
		\\
		\raisebox{0.8\totalheight}{\rotatebox[origin=c]{90}{GraphPillars}}
		& \picturewithcoordinates{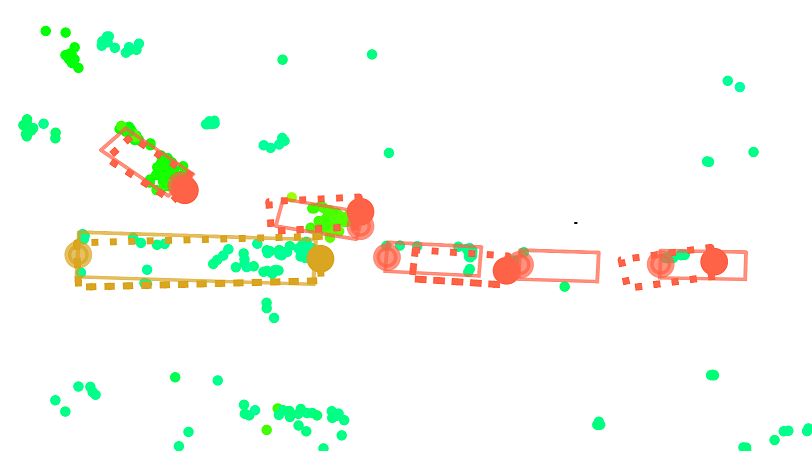}{(2.5, 0)}{(2.0, 0)}{(2.5, 0)}{(2.5, 1.0)}
		& \picturewithcoordinates{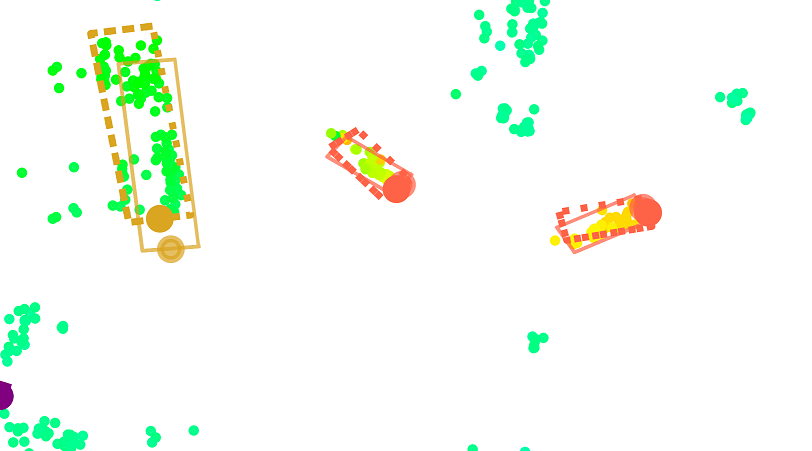}{(2.5, 0)}{(2.0, 0)}{(2.5, 0)}{(2.5, 1.0)}
		& \picturewithcoordinates{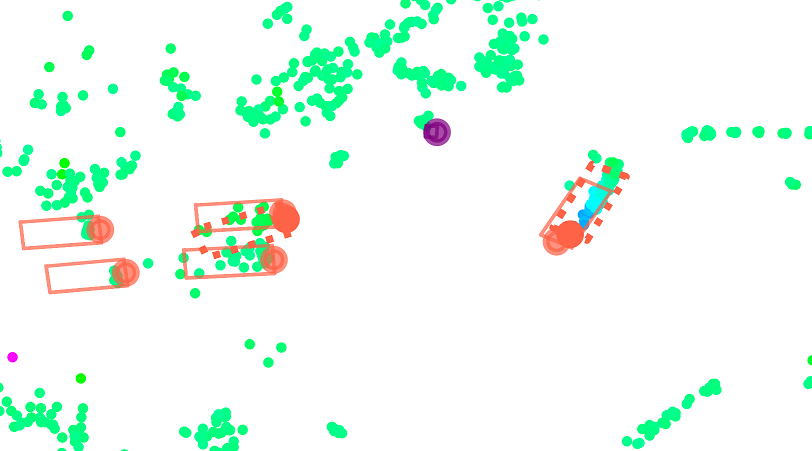}{(2.5, 0)}{(2.0, 0)}{(2.5, 0)}{(2.5, 1.0)}
		\\
		\raisebox{0.7\totalheight}{\rotatebox[origin=c]{90}{KPConvPillars}}
		& \picturewithcoordinates{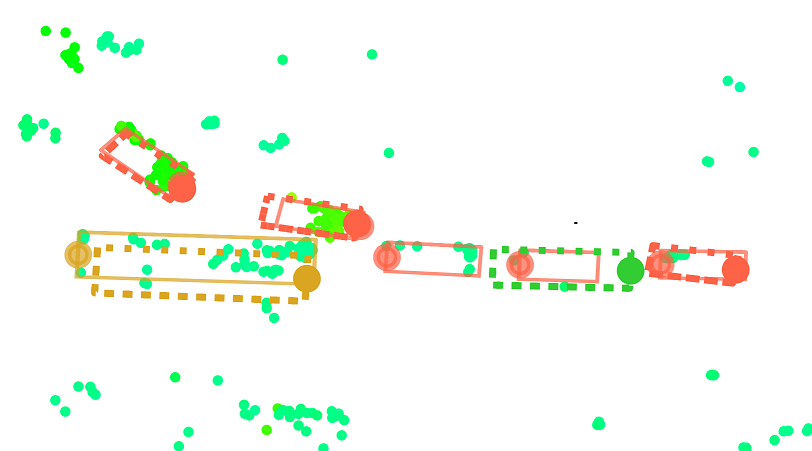}{(2.5, 0)}{(2.0, 0)}{(2.5, 0)}{(2.5, 1.0)}
		& \picturewithcoordinates{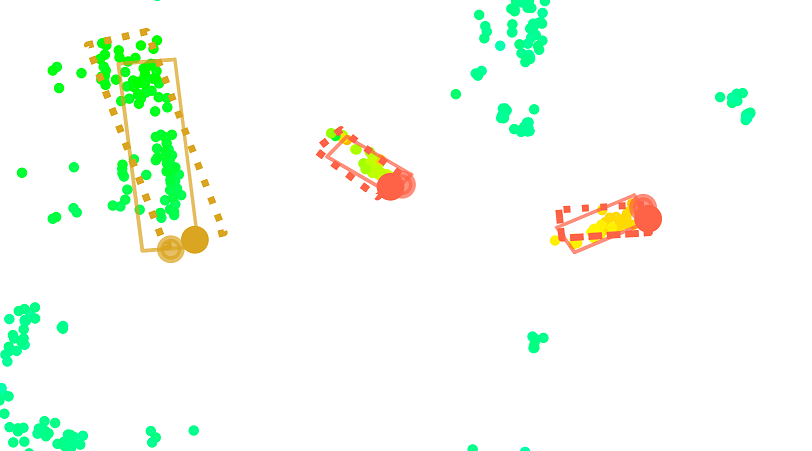}{(2.5, 0)}{(2.0, 0)}{(2.5, 0)}{(2.5, 1.0)}
		& \picturewithcoordinates{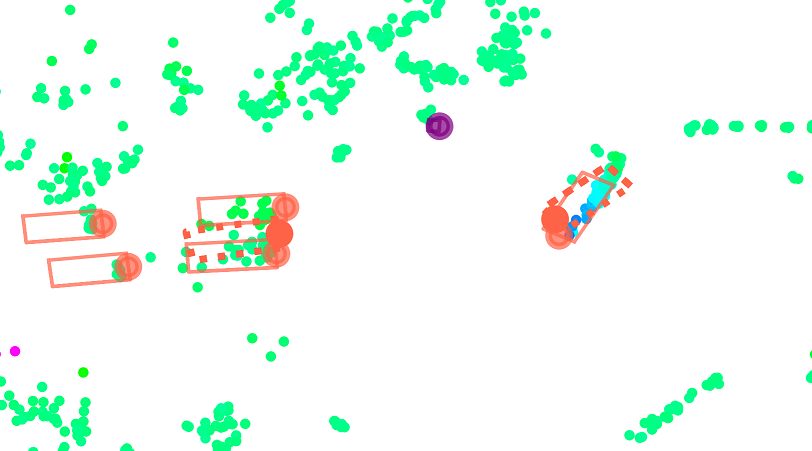}{(2.5, 0)}{(2.0, 0)}{(2.5, 0)}{(2.5, 1.0)}
	\end{tabular}
	\caption{
	Camera images and qualitative results for different samples on the nuScenes validation dataset. 
	Input points, detections and ground truth boxes are shown in a top-down bird's eye view. 
	Radar reflections are illustrated as color-encoded dots, where the color represents the ego motion compensated Doppler velocity and stationary targets are depicted in green. 
	Predicted boxes (dotted lines) and ground truth boxes (solid lines) are color-encoded according to the their class type. 
	Results are illustrated for two point-based (KPConv, GNN), one grid-based (PointPillar-like) and two of our novel hybrid architectures (GraphPillars, KPConvPillars). 
	It can be seen that the hybrid approaches clearly improve the orientation estimate and classification quality, especially for non-longitudinally oriented objects.  
	}\label{fig:exemplary_results}
\end{figure*}

\section{Conclusion}
To summarize, this paper presented a hybrid point-based and grid-based object detection network for automotive radar. 
The radar reflection point cloud is first processed using layers of a point-based object detection network to learn the local neighborhood context. 
After these layers, the features of the point cloud are rendered to a BEV grid, where they are further processed by a CNN. 
This approach combines the powerful detection performance of grid-based approaches with the continuous local neighborhood processing of point-based methods.
In experiments on the public nuScenes dataset, the hybrid models outperformed networks from previous literature and achieved a relative improvement for the detection performance by up to \SI{19.1}{\%} in mAP and for the orientation estimation by up to \SI{9.8}{\%} in AOE compared to solely grid-based methods.

%\addtolength{\textheight}{-12cm}   % This command serves to balance the column lengths
                                  % on the last page of the document manually. It shortens
                                  % the textheight of the last page by a suitable amount.
                                  % This command does not take effect until the next page
                                  % so it should come on the page before the last. Make
                                  % sure that you do not shorten the textheight too much.

%%%%%%%%%%%%%%%%%%%%%%%%%%%%%%%%%%%%%%%%%%%%%%%%%%%%%%%%%%%%%%%%%%%%%%%%%%%%%%%%

\bibliographystyle{ieeetr}
\bibliography{references}

\end{document}